\documentclass{article}

\usepackage{arxiv}

\usepackage[utf8]{inputenc} 
\usepackage[T1]{fontenc}    
\usepackage{hyperref}       
\usepackage{url}            
\usepackage{booktabs}       
\usepackage{amsfonts}       
\usepackage{nicefrac}       
\usepackage{microtype}      
\usepackage{lipsum}
\usepackage{graphicx}
\usepackage{xspace}
\usepackage{tikz}
\usepackage{amsmath}

\newcommand*\circled[1]{\tikz[baseline=(char.base)]{\node[shape=circle,draw,inner sep=1pt] (char) {#1};}}

\def\bolds{\ensuremath{\mathbf{s}}}

\newcommand\IGNORE[1]{\xspace}

\newcommand{\kaan}[1]{\textcolor{Blue}{\textbf{[Kaan: #1]}}}
\newcommand{\jan}[1]{\textcolor{Green}{\textbf{[Jan: #1]}}}

\title{Gaze-Sensing LEDs for Head Mounted Displays}

\author{
  Kaan Ak\c{s}it\thanks{Website: \texttt{kaanaksit.com}} \\
  NVIDIA\\
  Santa Clara, CA 95050 \\
  \texttt{kaksit@nvidia.com} \\
   \And
  Jan Kautz \\
  NVIDIA\\
  Santa Clara, CA 95050 \\
  \texttt{jkautz@nvidia.com} \\
   \And
  David Luebke \\
  NVIDIA\\
  Santa Clara, CA 95050 \\
  \texttt{dluebke@nvidia.com} \\  
}

\begin{document}
\maketitle

\begin{abstract}
We introduce a new gaze tracker for Head Mounted Displays (HMDs). We modify two
off-the-shelf HMDs to be gaze-aware using Light Emitting Diodes (LEDs). 
Our key contribution is to exploit the sensing capability of LEDs to create 
low-power gaze tracker for virtual reality (VR) applications. 
This yields a simple approach using minimal hardware to achieve good accuracy 
and low latency using light-weight supervised Gaussian Process Regression (GPR) 
running on a mobile device. 
With our hardware, we show that Minkowski distance measure based GPR
implementation outperforms the commonly used radial basis function-based
support vector regression (SVR) without the need to precisely determine free
parameters. We show that our gaze estimation method does not require complex
dimension reduction techniques, feature extraction, or distortion corrections
due to off-axis optical paths. We demonstrate two complete HMD prototypes with a
sample eye-tracked application, and
report on a series of subjective tests using our prototypes.
\end{abstract}

\keywords{gaze trackers; supervised learning; light emitting diodes}

\begin{figure}[htbp]
\centerline{\includegraphics[width=1.0\columnwidth]{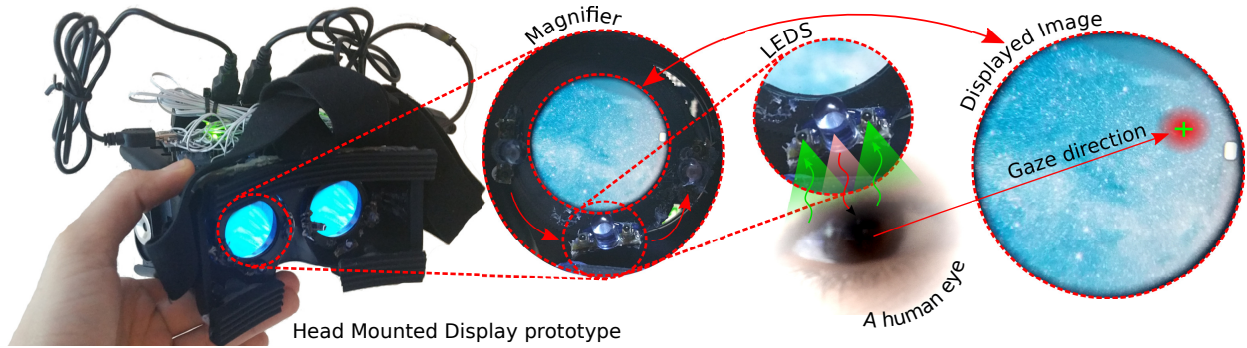}} 
\caption{Adding gaze tracking to Head Mounted Displays (HMDs) by exploiting sensing capability of Light Emitting Diodes (LEDs): An off-the-shelf HMD system is modified to demonstrate gaze tracking support only with LEDs (left). LEDs are placed in front of a human subject's eyes for sensing and illumination. The prototype comprises only a smartphone, two microcontrollers, and a low number of LEDs with a supervised adaptive pattern recognition algorithm. The overall system estimates the gaze direction of a human subject in real-time over a smartphone screen seen through a pair of magnifier lenses.} 
\label{fig:teaser}
\end{figure}

\section{Introduction}
\label{section:Introduction}
Head Mounted Displays use a variety of sensors to provide immersive interaction with engaging virtual reality experiences.  Emerging consumer HMDs for VR 
use gyroscopes, accelerometers, various optical sensors, and so on, either embedded in the headset or grouped into an external unit. These sensors track head orientation, user motions, whether or not a user is wearing the device, to provide user controls, and in short to enhance the user's experience in a virtual world.

Using gaze as an input modality \cite{jacob2003eye,majaranta2014eye} can be natural, fast, and has the potential to enhance the user experience in an HMD system. While decoupled non-mouse input modalities as a pointing mechanism for large displays has proven awkward, 
gaze-supported target acquisition has been found fast and natural~\cite{san2010gaze,stellmach2012look,turner2013eye,turner2014cross}. Hence, industry has been pushing towards employing gaze tracking as a key part of future HMD systems. Current prototype solutions are typically expensive (e.g., FoveVR\IGNORE{\footnote{http://www.getfove.com/}}, Tobii-Starbreeze\IGNORE{\footnote{http://www.tobii.com/group/news-media/press-releases/tobii-announces-eye-tracking-vrinitiative-andcollaborationwithstarbreeze/}}, SMI's HMD upgrade for Oculus Rift\IGNORE{\footnote{http://www.smivision.com/en/gaze-and-eye-tracking-systems/products/eye-tracking-hmd-upgrade.html}}), 
but there is also a growing interest towards low-cost gaze trackers with some promising results~\cite{san2009low,san2010evaluation,johansen2011low,mantiuk2012yourself}. 

We believe major challenges remain for conventional gaze tracking hardware and software:

\begin{enumerate}
\item Conventional gaze trackers rely on imaging techniques, which have relatively high power demands and may not be truly suitable for low-power mobile solutions; 
\item Imaging
 equipment typically introduces complexity in the software, adding an extra image processing block in the pipeline; and,
\item Conventional gaze trackers work with high-dimensional inputs (high-resolution images) and thus introduce latency at multiple stages, including the capture hardware, communication protocols, and image processing-based gaze estimation pipeline.
\end{enumerate}

We are motivated to address these issues by decreasing hardware and software complexity, seeking the simplest possible solution that provides useful-quality eye tracking. We use Light Emitting Diodes (LEDs) as the core of our gaze tracking hardware design.  Commonly used as illumination devices, LEDs in fact have well-known bi-directional characteristics \cite{dietz2003very,round1907note} with the ability to sense and emit light. Since they are also inexpensive, can perform color selective sensing, can illuminate and sense from the same physical location, and are easily controlled without a dedicated custom circuitry, LEDs constitute a good choice as a hardware solution.


In this paper, we describe how to take advantage of LED's bi-directional characteristics to allow both light capture and illumination. We position an array of LEDs in front of a human subject's eyes for the task of gaze estimation, in the context of an HMD system. As a human subject observes a scene projected at a fixed virtual plane, we illuminate the subject's eye with infrared (IR) light from different perspectives in a time multiplexed fashion. We capture and digitize the intensities of IR light reflecting off the eye from different perspectives with high refresh rates. Using our supervised adaptive pattern recognition implementation, we accurately estimate the gaze location of the human subject at the virtual image plane.

\subsection{Contributions}
\label{subsection:contributions}

\textbf{All-in-one Sensor Technology:} The core item in our design, an LED, can be used as an illuminator or as a sensor for integrating light intensity over a certain field of view (FOV) and exposure interval. To our knowledge, this is the first time that LEDs are used in a gaze tracker both for capture and illumination.

\textbf{Less Hardware and Lower Cost:} Our design employs fewer components. To our knowledge, our design has the simplest electronics design yet demonstrated. We believe that our hardware design leads the category of ultra-low cost gaze tracker. While decreasing cost, we show that accuracy and sampling rates can match existing devices.

\textbf{Supervised Adaptive Gaze Estimation Algorithm:} We use a supervised adaptive pattern recognition algorithm with our sensor technology. We show that our method has low computational demands and runs fast enough on a conventional mobile platform. We compare against the most common methods from the literature, and demonstrate equivalent or better accuracy with a much simpler hardware.

\textbf{Complete Prototype:} We demonstrate two different complete custom HMD systems, created by modifying off-the-shelf items. We also demonstrate example eye tracked applications with our prototype, and provide results of an informal subjective test with human subjects.

\subsection{Benefits and Limitations}
\label{subsection:benefitsandlimitations}

\textbf{Benefits:} 
Today's HMD manufacturers are competing to create the least bulky HMD. One way to achieve this goal is to reduce the bulk caused by each component of the system. Off-the-shelf LEDs are lighter and smaller than off-the-shelf cameras. Thus, our proposed method uses less volume and less weight than an alternative using single or multiple camera(s). 

LEDs also offer an advantage in power consumption over camera based solutions. Possible heat dissipation related issues caused by cameras are also avoided with our low power consuming system. Additionally, our gaze estimation pipeline is computationally light-weight. Thus, our proposed technique is a good match for battery operated applications.
Since we only employ a small number of sensors, compared to camera-based solutions (which have millions of sensing pixels), latency within the hardware and software is reduced and results in higher sensing rates. 
\IGNORE{For instance, we do not require a complex dimension reduction algorithm such as principal component analysis (PCA), or feature extraction algorithm such as edge detectors.
\jan{I don't understand what the point the following sentence is trying to make.}\kaan{Moved to discussion.} 
}

\IGNORE{Most of the gaze tracker hardware that use cameras already employ an illumination source, typically an LED. We believe our pipeline can be a part of hybrid system, in which there are already LEDs used as illumination sources for cameras.\jan{Would move this to a discussion section. Detracts here.}\kaan{Moved to discussion.}}


Our method offers good accuracy. We show that mean angular error can be as low as $1.1^o$ with a median angular error of $0.7^o$. 

\textbf{Limitations:} While providing sufficient accuracy for a variety of VR applications, due to their poor sensing characteristics compared to photodiodes LEDs may not be suitable choice for applications requiring very high accuracy ($error < 0.5^o$), such as in psychological, neurological and reading research \cite{holmqvist2012eye}. 

Our proposal requires a larger amount of calibration data than conventional imaging based gaze tracking systems, thus the initial phase of calibration is comparably longer. 

Conventional HMDs can shift slightly on a subject's face, commonly due to fast head movements. In turn, this causes the sensing hardware to shift relative to the subject's eyes. 
Using the original calibration data makes gaze estimates less reliable. 
We share this common problem with other gaze trackers. We choose to recalibrate in such cases, which can be burdensome to users. 

\section{Related Work}
\label{section:RelatedWork}

\textbf{	Sensing with LEDs:} LEDs are known to be reliable light sensors, and have been used in other bi-directional systems such as visible light communication (VLC) systems  \cite{6852086,schmid2014visible}, temperature and pressure sensing systems \cite{lange2011multicolour}, bidirectional reflectance distribution function (BRDF) measurement systems \cite{ben2008led}, color sensing and illumination mimicking applications \cite{li2014color,li2015led}, and human shape sensing with VLC \cite{li2015human,an2015visible}. 

To our knowledge, this is the first time LEDs are used as sensors in a gaze tracker application.

\textbf{Gaze Tracker Hardware:} 
The foundation of any gaze tracker is the hardware used to capture the raw data.  The "scleral search coil," for instance, requires the user to wear a contact lens with copper coils on top while exposed to an alternating magnetic field~\cite{robinson1963method}. The scleral search coil is reported to provide $15$ arc seconds of angular resolution with $1000~Hz$ sampling capabilities. Despite this impressive accuracy, the scleral search coil fails to to provide a comfortable user experience and is not an option for consumer products. Another common methodology is electro-oculography (EOG) \cite{marg1951development}, which uses electrodes placed around the eye along the horizontal and the vertical axes. However, the technique is generally known to suffer from resolution limitations, and shares the 
 same discomfort problem with a scleral search coil.

Gaze trackers using light are categorized as imaging and non-imaging optical devices.  Relatively few imaging gaze trackers have explicitly tackled the problem related to power-consumption and computation-intensity: iShadow \cite{mayberry2014ishadow}, an imaging system, reported a 70 mW power consumption and $30~Hz$ estimation rate with an angular resolution of $3$ degrees. OLED-on-CMOS technology \cite{SDTP:SDTP1540} has 
been demonstrated for capturing an eye's images from a large spectrum of light ($600-900~nm$) and displaying those images, however, 
in its current state, the technology's display resolution is very limited, and negatively effecting a display's fill factor. 

Non-imaging gaze trackers provide another approach. Naugle and Hoskinson~\cite{naugle2013two} demonstrated two different methods using binary gaze tracking with a pair of LEDs and a photodiode for a low field-of-view (FOV) ($<20^o$) HMD. While this setup reduces the power consumption of the display by over $75~\%$, it only provides limited information on whether the user is actively wearing the display or not. 

More closely related to our proposal, Topal et al.~\cite{topal2014low} demonstrate a low-computational overhead, non-imaging gaze tracker based on IR light-emitting diodes and IR sensors around an eyeglasses frame. Another similar non-imaging gaze tracker~\cite{grover2005eye} contains a ring of LED emitters that are stimulated in sequence. Unlike our proposal, the tracker contains photodiode sensors, which are measured for each LED stimulation. These systems, however, require head pose to be completely fixed or stabilitized with a mechanism such as a bite-bar. In contrast, our prototype hardware does not depend on fixed head pose, does not use photodiodes nor sophisticated or custom light-sensing circuits, and is embedded in an HMD context without producing occlusion in front of the eye. Our algorithmic approach is also simpler and less demanding on the hardware, using Gaussian Regression Processes (GPR) \cite{Rasmussen06gaussianprocesses} for gaze estimation rather than a model-based linear or non-linear mapping --- thus avoiding tuning of algorithm parameters.  

We believe our solution provides the simplest hardware design yet for a gaze tracker embedded inside an HMD.

\textbf{Adaptive Gaze Estimation Algorithms:}  
Because each human subject introduces multiple differences for the input of a non-imaging gaze tracker, we believe that the task of gaze estimation, in this case, is a better match for an adaptive supervised pattern recognition technique rather than finding a good model for noisy sensor input.
Traditionally, supervision of such pattern recognition techniques in gaze trackers 
has been incorporated into a calibration procedure. 

Within the realm of imaging gaze trackers, the relevance vector regression (RVR)~\cite{6467271} method was found to be more effective than support vector regression (SVR). RVR shares the same functional form as SVR. Unlike SVR, however, RVR tries to find the weights of the regressor from training data. We have experimented with SVR with radial basis function (RBF) as the kernel, and we found that GPR with a non-parametric similarity measure provides an accurate result without the comptuational burden of finding the correct weights using RVR. This is important since finding correct weights online may not be a feasible task for limited processing power on mobile devices.

Noris et al.~\cite{noris2008calibration} show 
that imaging gaze trackers can benefit from GPR for calibration-free operation. Their implementation used principle component analysis (PCA) for feature dimension reduction. As our technique depends on a low number of sensory inputs, 
we did not find dimension reduction necessary for our gaze estimation algorithm, and 
instead we use GPR directly for the task of gaze estimation.


Using cost effective hardware (albeit with a relatively power hungry imaging gaze tracker), Sewell and Komogortsev~\cite{sewell2010real} demonstrated effective usage of a neural network for offline training, and demonstrated a relatively low error ($<3.68^o$). 
Our method achieves a much lower error, and we also demonstrate that in specific applications,
online training can be enabled.


\section{System Overview}
\label{section:systemoverview}

Gaze tracking devices for HMDs estimate where a user is gazing relative to a virtual image plane as seen through the HMD's lenses. The task of gaze estimation is a layered problem that requires the integration of sensors, optical design, image/signal processing, and electronics design. The following sections describe our approach in addressing each of these tasks.

\subsection{Sensing with LEDs}
\label{subsection:SensingWithLEDs}

At the core of our design, LEDs are used to capture light and to illuminate the eye. \IGNORE{LEDs are semiconductor materials with a p-n junction connected to two electrical terminals.} LEDs with infrared light emission are typically used in HMD gaze tracker hardware, since humans are insensitive to IR illumination~\cite{kunka2009non}. 
A human eye's cornea has similar absorption and reflection characteristics in the near IR as in visible light \cite{van1997near}. Furthermore, IR LEDs have a narrow bandwidth (typically \IGNORE{\footnote{http://www.adafruit.com/datasheets/IR333$\_$A$\_$datasheet.pdf}} $\sim 50~nm$), avoiding cross-talk with other wavelengths.

\begin{figure}[htbp]
\centerline{\includegraphics[width=0.5\columnwidth]{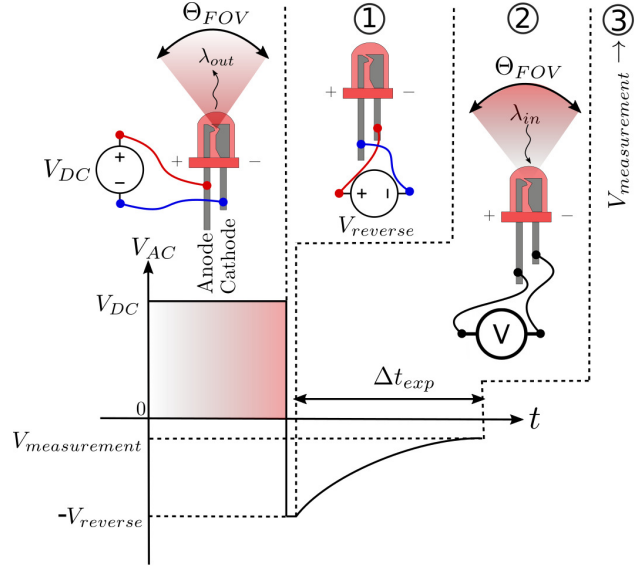}} 
\caption{Illustration of different modes of a bi-directional LED (from left to right): (1) applying a forward voltage of $V_{DC}$, in which the LED emits light with a wavelength of $\lambda_{out}$ and an emission cone angle, $\Theta_{FOV}$; (2) applying a reverse voltage pulse, $V_{reverse}$, for a short time duration, discharging LED with incoming light that has a wavelength of $\lambda_{in}$ for a specific time, $\Delta t_{exp}$, with an reception cone angle of $\Theta_{FOV}$; and (3) measuring a voltage, $V_{measurement}$, from the LED.}
\label{fig:BidirectionalLED}
\end{figure}

LEDs provide illumination when a forward voltage is applied to their two electrical terminals. However, LEDs can also act as photodetectors \cite{dietz2003very} 
by following three steps: \circled{1} Apply a reverse voltage pulse for a short time duration. \circled{2} Discharge LED's capacitance immediately afterwards. \circled{3} Measure the voltage across LED to determine how much discharge of capacitance took place after a certain time. Figure~\ref{fig:BidirectionalLED} illustrates each of the explained steps.
\IGNORE{\footnote{http://makezine.com/projects/make-36-boards/how-to-use-leds-to-detect-light/}}

These steps can be easily implemented by wiring the LED to a microcontroller for full control over charge, discharge, and measure. The microcontroller is controlled by and relays measurements to a host. Typically, LEDs are most sensitive to wavelengths $\lambda_{in}$ that are shorter than their emission spectrum (so $\lambda_{in}<\lambda_{out}$) \cite{lange2011multicolour}. 
Thus, larger exposure times are required if LEDs with the same emission spectrum are used. To achieve lowest possible latency with a given configuration, we choose to use different LEDs that have intersecting emission and sensing spectra in the IR range. 

Eye safety is also important when a user is exposed to infrared radiation; $\Delta t_{exp}$ and maximum irradiance of an LED must be considered according to safety regulations for infrared light sources \cite{boucouvalas1996iec}. 

\begin{figure}[htbp]
\centerline{\includegraphics[width=0.5\columnwidth]{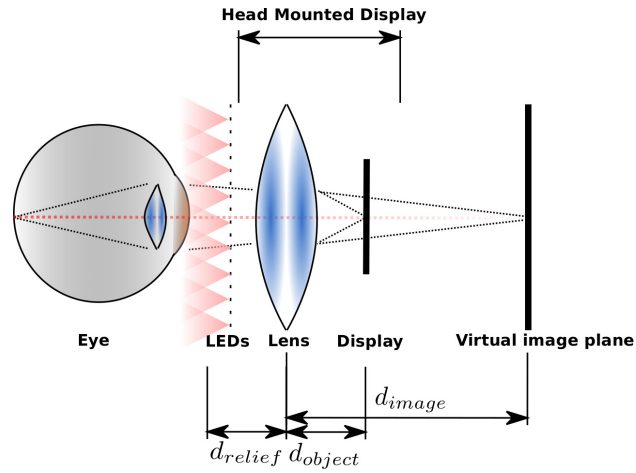}} 
\caption{A sketch showing our proposed configuration within a conventional HMD.}
\label{fig:optics}
\end{figure}

\textbf{Positioning of LEDs:} To use LEDs for capture and illumination, LEDs must be placed at specific locations in front of an eye, or optical path must be relayed to have the same effect.  Currently the most common configuration for the optics of a commercial HMD, as shown in Figure \ref{fig:optics}, uses a pair of magnifier lenses placed in front of a display to create a virtual image at some distance in front of a user. Such an HMD setting typically includes a distance from eye, called \emph{eye relief} $d_{relief}$ of $25-30~mm$, and an additional spacing $d_{object}$ of $35-50~mm$ between the magnifier lens and display, which is determined by the focal length of the magnifier lens. This leaves two obvious options of where to place the LEDs -- between the lens and the eye or between the magnifier lens and the display.


From the stand point of integrating our solution, the easiest place to put the LEDs would be directly in front of the user's eyes, as shown in Figure~\ref{fig:optics}. This arrangement also has the advantage that it minimizes light loss due to scattering off the other optical elements of the HMD.
However, LEDs positioned on-axis in front of an eye would occlude parts of the image. Placing LEDs in an off-axis arrangement avoids such occlusions. 

\begin{figure*}[htbp]
\centerline{\includegraphics[width=1.\columnwidth]{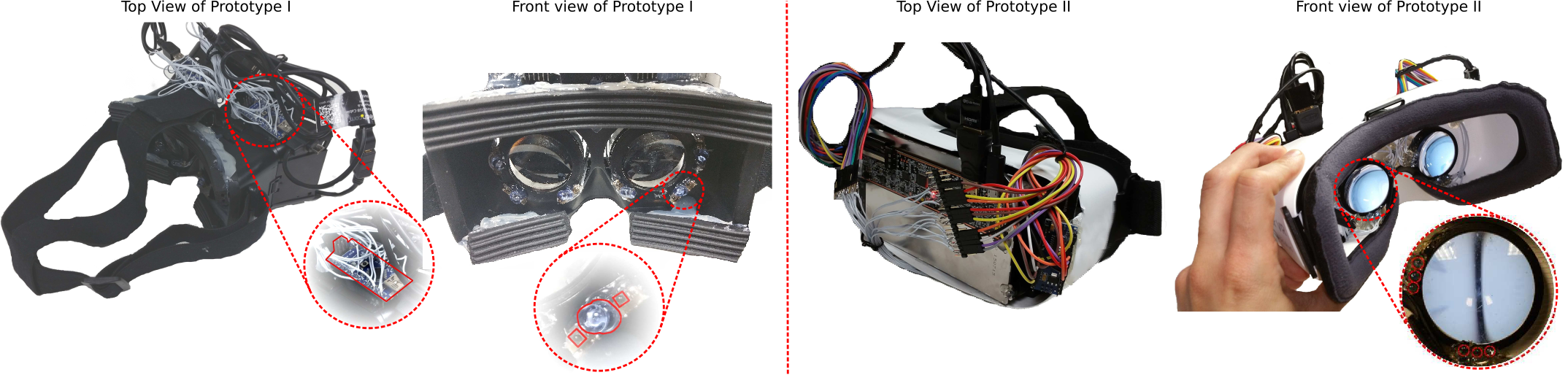}} 
\caption{Left: A pair of photographs showing top and front view of our first prototype. Zoomed region in top view shows a microcontroller. Zoomed region in front view shows three LEDs, center one that is highlighted in red is used in illumination source mode, and neighbouring two that are highlighted in green are in receiver mode in our driving scenario. Note that all LEDs in our prototype can switch to both modes upon request. Right: A pair of photographs showing top and front view of our second prototype. Zoomed region in front view highlights six LEDs that are used for both sensing and illumination.}
\label{fig:prototype}
\end{figure*}

%
We rely on the experimental study from Nguyen et al.~\cite{nguyen2002differences} to determine the best positions. A maximal amount of corneal and pupil reflection can be achieved, when the sensing LEDs are close to the illuminating LEDs. We thus place LEDs side-by-side in a ring around the magnifier lens. Front view in Figure \ref{fig:prototype} shows a layout for positioning of LEDs in our prototype. We arrange the LEDs as groups of two sensing LEDs with an illuminating LED in between. 

\subsection{Estimation Using Captured Data}
\label{subsection:datacapture}

\begin{figure}[htbp]
\centerline{\includegraphics[width=0.5\columnwidth]{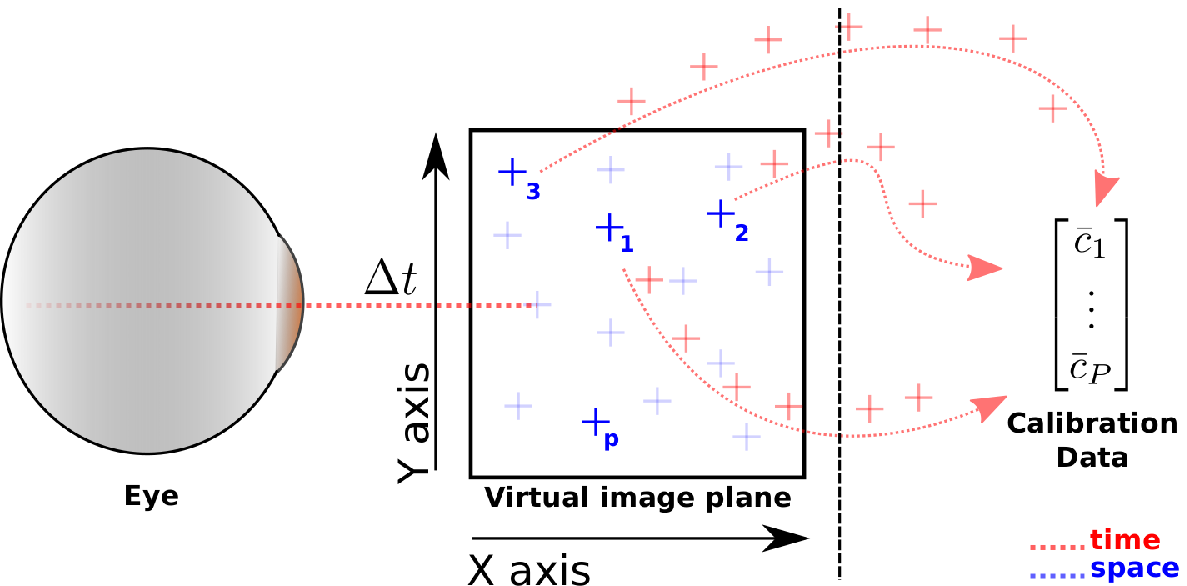}} 
\caption{A sketch showing construction of a calibration matrix in a simplified way. A user gazes 
at a certain point at a virtual image plane for a time period of $\Delta t_{\mathrm{fix}}$. During that time data is captured, and the arithmetic mean value of the capture, $\bar{c_p}$ is stored at a corresponding row of the calibration matrix.}
\label{fig:CalibrationMatrix}
\end{figure}

Our capture hardware hosts a fixed number $M$ of sensing LEDs. The capture hardware transmits a data capture vector to the host (PC, smartphones, and etc.). This vector contains measured data as
\begin{equation}
\bolds(t) = [s_0(t), s_1(t), \dots, s_{M}(t)],
\label{equ:vector}
\end{equation}
where each $s_m(t)$ represents the output of the $m$-th LED connected to the capture hardware. We construct a calibration matrix that represents the relation between a captured vector and the gaze locations as follows. First, pre-defined locations are highlighted on the virtual image plane in random order, and users are asked to dwell on each of the highlighted locations for a certain duration $\Delta t_{\mathrm{fix}}$. During this time, we sample $\bolds$ multiple times at fixed intervals $\Delta t_{\mathrm{v}}$, and store the mean of the measurements as $\bar{c}_p = \sum_t \bolds(t)$ for each predefined location $p$. To ensure a meaningful calibration, we check that the variance of all the measurements for a location $p$ is below a certain threshold, otherwise we discard the measurements for that location.
Figure~\ref{fig:CalibrationMatrix} summarizes the procedure for creation of a calibration matrix.

The pre-defined calibration points can be at any location. 
Typically, we use a grid of $2\times2$ to $5\times5$ evenly spaced locations. Starting from this small number of calibration points, we need to derive the gaze location given a set of new measurements $\bolds(t)$. At a later stage, calibration data is enlarged online through a gaming application, in which a user's task is to gaze and destroy opponents. Using collected data, we have evaluated two different regression methods: support vector regression (commonly used in prior research), and gaussian process regression (our choice).

\textbf{Support Vector Regression:} SVR has a generalized form as
\begin{equation}
\begin{aligned}
\begin{bmatrix}
e_x \\
e_y
\end{bmatrix}
=k^T
\begin{bmatrix}
u_x \\
u_y
\end{bmatrix}^T,
\end{aligned}
\label{equ:SupportVectorRegression}
\end{equation}
where $e_x$ and $e_y$ represents estimated gaze position along $x$ and $y$, $k^T$ represents a vector that contains the similarity measures between the captured $\bolds(t)$, and the calibration vectors $\bar{c_p}$. Finally, $u_x$ and $u_y$ represent vectors that correspond to the $x$ and $y$ position of each $\bar{c_p}$.

A regression technique depends on a distance measure, that indicates how similar the captured data is to the stored calibration data. Such a comparison using any method will provide a similarity vector as in
\begin{equation}
k=
\begin{bmatrix}
\kappa(\bolds(t),\bar{c_1}) \\
\vdots            \\
\kappa(\bolds(t),\bar{c_P}) \\
\end{bmatrix},
\label{equ:SimilarityVector}
\end{equation}
where $\kappa(a,b)$ donates the used distance function to determine similarity in between vector $a$ and $b$, $k$ represents the distance vector. Our choice of distance measure for SVR calculations is the commonly used radial basis function (RBF) \cite{6467271,noris2008calibration}. 

\textbf{Gaussian Process Regression:} Through our experiments, we found GPR to be a robust and accurate regression method. GPR takes the following general form:
\begin{equation}
\begin{aligned}
\begin{bmatrix}
e_x \\
e_y
\end{bmatrix}
=k^T C^{-1} 
\begin{bmatrix}
u_x \\
u_y
\end{bmatrix}^T,
\end{aligned}
\label{equ:GaussianProcessRegression}
\end{equation}
with variables as described above.
The covariance matrix $C$ is calculated as 
\begin{equation}
C=
\begin{bmatrix}
\kappa(\bar{c_0},\bar{c_0}) & \dots  & \kappa(\bar{c_0},\bar{c_p})\\
\vdots                 &                        &        &                  \\
\kappa(\bar{c_p},\bar{c_0}) & \dots  & \kappa(\bar{c_p},\bar{c_p})\\
\end{bmatrix}.
\label{equ:CoverianceMatrix}
\end{equation}
%
%

\textbf{Distance Measures:} Comparing a vector with another vector can be accomplished in multiple ways. Although, we have evaluated multiple different distance measures (\textit{Cosine, Minkowski, Manhattan, Canberra})  \cite{bray1957ordination,lance1966computer,Rasmussen06gaussianprocesses}, we found the Minkowski distance measure to be the most effective to be used with the GPR algorithm:
\begin{equation}
\kappa(a,b) = \left( \sum_{i=1}^{n} w_i\left| a_i - b_i \right|^m \right)^{1/m},
\label{equ:WeightedPNorm}
\end{equation}
where $a$ and $b$ are two vectors to be compared, $w_i$ is the weighting factor for a certain channel in the captured data, $m$ is the degree of the norm, and $i$ is the index of the element.
We find $m=2$ and setting $w_i=1$ to yield good results; more details will be
presented in Section~\ref{section:evaluation}. 
 
For the SVR algorithm, we employed the RBF kernel:
\begin{equation}
\kappa(a,b)= e^{-\frac{\mid\mid a- b \mid\mid}{2\sigma ^2}},
\label{equ:RBF}
\end{equation}
in which $\sigma$ represents a free parameter. We use a grid-search to find the optimal $\sigma$. 

\section{Implementation}
\label{section:implementation}
\label{subsection:Prototype}

Two different off-the-shelf HMDs are transformed into a gaze sensing HMD using our methodology. In this section, we introduce both design choices with practical aspects. Our prototypes are shown in Figure \ref{fig:prototype}. 

Our first prototype consists of 3 LEDs\IGNORE{\footnote{http://www.adafruit.com/product/388}} per eye functioning as light sources, 6 LEDs\IGNORE{\footnote{http://www.osram-os.com/Graphics/XPic8/00082747$\_$0.pdf}} per eye functioning as light sensors, a smartphone, \IGNORE{\footnote{http://www.samsung.com/global/microsite/galaxynote4/}} an Arduino Nano\IGNORE{\footnote{https://www.arduino.cc/en/Main/ArduinoBoardNano}} microcontroller (uC) per eye, 
a controller\IGNORE{\footnote{http://gaming.logitech.com/en-us/product/f710-wireless-gamepad}}, and a VR headset\IGNORE{\footnote{http://sunny-peak.com}} as a housing. Our gaze tracking algorithm runs on a smartphone in synchronism with two uCs driving LEDs.

Our second prototype consists of 6 LEDs per eye functioning as both light sensors and light sources, an Arduino Nano\IGNORE{\footnote{https://www.arduino.cc/en/Main/ArduinoBoardNano}} microcontroller (uC) per eye,  a HDMI supported 2K resolution display, a controller, and a VR headset as housing. This time, our gaze tracking algorithm runs on a desktop computer.

\textbf{Optics:} Our both prototypes use a pair of magnifier lenses in front of a display, as shown in Figure~\ref{fig:optics}. 
Considering the magnification of the lens, and the distances between LCD, lenses, and eyes, and the display's pixel density, each pixel corresponds to a visual angle of 0.12 degrees in our both prototypes.
%
%
%

\textbf{Sensing Electronics:} LEDs are hooked to the two uCs for both prototypes, so that they can be programmed accordingly. In the case of our first prototype, illuminator LEDs are attached to digital input/output (IO) pins, and the sensing LEDs' anodes were attached to digital IOs, and their cathodes were attached to Analog-to-Digital (ADC) pins of the uC. In the case of our second prototype, LEDs are only attached in the same way as the sensing LEDs in our first prototype. Each time a LED is requested to sense, it follows the pattern of \circled{1}-\circled{2}-\circled{3} from Section~\ref{subsection:SensingWithLEDs} and Figure~\ref{fig:BidirectionalLED}. LEDs have a soft-coded mechanism that adjust $\Delta t_{exp}$ per LED basis, so that saturation caused by varying light conditions can be avoided for each LED.  


Identical two LEDs typically have a limited overlap in their emission-sensing spectra. Thus, leading to low resolution capture with less sampling frequency when identical LEDs are used. In our first prototype, we choose to dedicate two different LEDs to specific tasks to maintain good resolution. However, with our second prototype, we also evaluated a configuration that uses the identical LEDs for illumination and sensing.

In our first prototype, an illuminator LED is turned on shortly before taking a measurement from its pair of neighbouring sensing LEDs. In our second prototype, all the remaining LEDs are illuminating the scene at the time of a capture from a LED. In our prototypes, used uC only allows a time multiplexed capture routine -one capture from a single LED at a time-. However, a simultaneous capture from different LEDs can be achieved with a different uC that has discrete ADCs for each analog pin. Thus, the effective sampling rate would increase significantly and latency would reduce even further. 

\textbf{User-interface application:} The uC works hand-in-hand with the user-interface application using a USB connection.
%
Our application deals with a number of predefined tasks: (1) collecting measurements from each LED by requesting them from the two uCs, (2) updating the user interface, (3) producing the actual gaze estimation using GPR, and (4) keeping logs related to captured data (event time stamps, and so forth).


\begin{figure*}[htbp]
\centerline{\includegraphics[width=1\columnwidth]{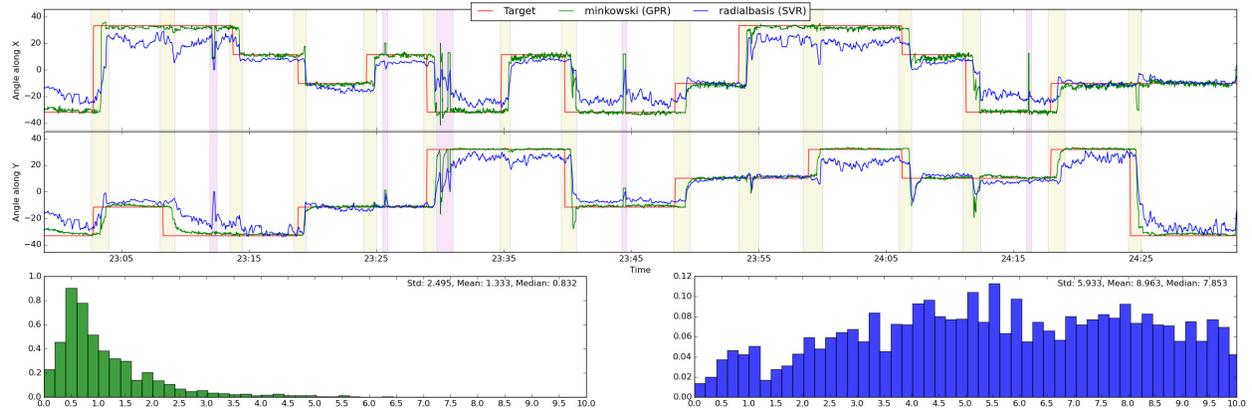}} 
\caption{Top row shows angular positions of a target on a screen along X axis in red color among with the outputs of GPR with Minkowski and SVR with RBF in green and blue color respectively. Middle row shows angular positions of a target on a screen  along Y axis in red color among with the outputs of GPR with Minkowski and SVR with RBF in green and blue color respectively. Both in top and middle rows, Regions highlighted with Magenta color shows blink events, and regions highlighted with yellow color shows Saccadic Reaction Time (SRT) with a gaze transition (saccade). Bottom row shows normalized histograms of angular errors for GPR with Minkowski and SVR with RBF in green and blue color respectively. There are $82$ calibration points in this sample dataset ($16$ from calibration grid, $66$ from VR gaming). Captured data used in making is from our first prototype.}
\label{fig:accuracy}
\end{figure*}

\section{Evaluation}
\label{section:evaluation}

Success of a regression algorithm for the task of gaze estimation is heavily affected by a number of factors, such as noise levels in the captured signal, positioning of LEDs in front of an eye, the used distance measures, and the number of stored points in a calibration matrix. We evaluate our method with respect to each factor. 

Our first prototype uses 18 illuminating LED and 12 sensing LED in total. Our first prototype is supplied by a voltage source with galvanic isolation (smartphone's battery), thus leading to more reliable analogue to digital conversions. On the other hand, our second prototype uses 12 LEDs in total for both illumination and capture. Our second prototype interfaces with a personal computer that is typically known to suffer more from noise caused by electromechanical parts, power-supply and supply-line. Noises can propagate through data and power lines. Thus, it requires more levels of filtering in circuitry and in digital means, which are causing latency both in digital estimation and analogue signal. We try to tackle noise problems at most with an additional layer of first order Infinite Impulse Response (IIR) digital low-pass filters, and a physical USB isolator. At extreme, a custom dedicated circuitry for analogue to digital conversions can outperform other options.

We report our findings on accuracy, and number of calibration data using our first prototype. We extend our subjective tests using our second prototype. Thus, we also report on practicality, and inter-intra personal differences using the second prototype. We also share our findings on sampling rates of both prototypes.
%
Our evaluation method depends on the experimental procedure summarized in Table~\ref{table:procedure}. We have conducted a series of subjective tests using this procedure, and recorded each session for analysis.

\label{subsection:ExperimentationProcedure}
\begin{table}[htbp]
\small
\begin{center}
\begin{tabular}{ p{3mm} p{73mm} }
  \hline
  \multicolumn{2}{l}{Instructions, levels and hints for a session}\\
  \hline 
   & \\
  L1 & User wears the headset.\\
  L2 & User starts the application by pressing "X" button in the controller.\\
  L3 & User is verbally asked about image quality. If not satisfactory, session is terminated. Otherwise user continues with the session. \\  
  L4 & Application shows a visual containing a rabbit character together with a text that states press "X" button and follow the rabbit.\\
  L5 & Rabbit on the screen stops after a while, a text in the visual appears, which commands a user to gaze at the rabbit constantly until it disappears. \\
  L6 & Another visual appears with an instruction, which commands a user to gaze at a red dot constantly, press "X" button on controller, and gaze constantly until red dot turns to green. Once it is green, another red dot appears at a different location, and user repeat his previous actions until there is no more red dot in the screen.\\
  L7 & User follows the instructions from L6 for each event objects.\\
  L8 & Another visual appears with a new instruction, which states that there will be a ghost character among with multiple characters in the next screen. Task is to destroy ghost, the user has to gaze at it for a certain time period. If user is gazing at the ghost, the ghost turns blue or other objects are changing colors; the user is asked to keep on gazing at the ghost and press "X" button on the controller.\\
  L9 & As the user is gazing at each object, the calibration data is augmented with the new incoming measurements from each failure case.\\
   & \\  
  \hline
\end{tabular}
\end{center}
\caption{Our experimental procedure, including the guidance provided to our participants.}
\label{table:procedure}
\end{table}




\textbf{Accuracy:} The accuracy of a gaze tracker system refers to the difference between the true and the measured gaze direction. Given a set of sample input data, we estimate the gaze locations using both GPR with Minkowski distances and SVR with RBF as the regressor. We show the angular locations of the target gaze locations along with the estimates from both methods in Figure~\ref{fig:accuracy}. Our error metric here is visual angular errors, which is calculated using Euclidean distances in between output of estimations and gaze targets. 
Figure~\ref{fig:accuracy} contains histograms of visual angular errors recorded over a fixed time duration for both techniques using a set of captured data from our first prototype. We have conducted a comparison of multiple different distance measures using GPR and SVR, a detailed analysis of these can be found in the supplementary material. 

We point out that regions highlighted with magenta color in Figure~\ref{fig:accuracy} show blinks, which lead the estimated gaze direction to have sharp jumps. Regions highlighted with yellow color in Figure~\ref{fig:accuracy} show regions, where a gaze target changes position in space. Humans are known to have a response time before an event of gaze change (Saccade) \cite{mazumdar2014comparison}, which is known as Saccadic Reaction Time (SRT). SRT may vary from individual to individual, affected by biological factors such as age and health conditions. Angular errors have been calculated by excluding highlighted regions. We have performed a user study among $5$ subjects with varying age, and eye prescription. Using GPR with Minkowski distances, we report our findings for each subject in terms of accuracy in Table \ref{table:accuracy}.

\begin{table}[htbp]
\begin{center}
\begin{tabular}{ c c c c }
  \hline
  & Mean & Median & Standard Deviation\\
  \hline
  Subject 1  & 1.34 & 0.83 & 2.49\\
  Subject 2  & 2.10 & 1.23 & 2.80\\      
  Subject 3  & 1.90 & 1.84 & 1.43\\
  Subject 4  & 1.40 & 1.04 & 1.10\\
  Subject 5  & 1.10 & 0.66 & 2.14\\   
  \hline
  Total Mean & 1.57 & 1.12 & 2.00\\
  \hline
\end{tabular}
\end{center}
\caption{Mean, median, and Standard Deviation values of each subjects using data collected from each subject. Angular errors are reported only for the case of using GPR with Minkowski distances.}
\label{table:accuracy}
\end{table}

Commercial gaze trackers (SMI, FOVE VR, etc.) for HMDs claim to have an accuracy ranging from $0.2^o-1.5^o$ in a controlled laboratory environment. We show that our mean angular accuracy can be as low as $1.10^o$ with a much simpler hardware setup in casual use case. We believe that our approach is a promising alternative to existing methods. 

\textbf{Number of calibration points and LEDs:} The number of calibration points and LEDs are other important variables in our design space. Thus, we investigate the resulting angular error when varying their number. Given a small number of calibration points, we have started our evaluation using only four channels with the highest variation in digital signals and have iteratively increased the number of channels. We have repeated this evaluation utilizing an increasing number of calibration points. 
%
%
Our findings suggest that smaller numbers of LEDs can produce similar angular accuracy levels. Adding more LEDs seem to only marginally enhance accuracy. However, we would like to highlight that an LED's contribution to the gaze estimate depends on its physical location and direction relative to a particular subject. I.e., depending on a subject's physiognomy
 (smaller vs. larger eyes, distance between eyes, and so forth), a particular sensing LED may contribute more or less. Thus, a larger number of LEDs compensate for such differences and add robustness. 
%

\begin{figure}[htbp]
\centerline{\includegraphics[width=0.5\columnwidth]{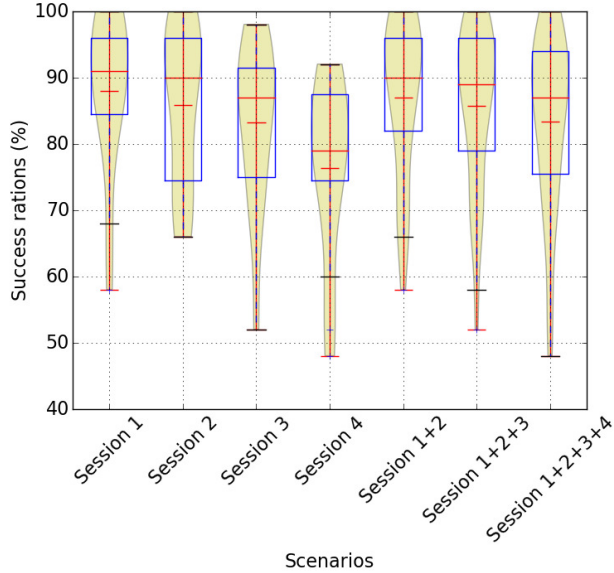}} 
\caption{Violin-box diagrams showing distribution of success rations from different scenarios. Small and large red vertical lines indicates mean, and median values respectively. Dashed bars indicates 1.5 times interquartile range (IQR), and the blue boxes indicates first and third quartiles from a distribution.}
\label{fig:statistics}
\end{figure}

\textbf{Practicality and inter-intra personal differences:} We extended our experiments using our second prototype. Total number of independent different subjects are 10 males and 4 females. At each experiment, a subject has to complete four independent sessions, in which a user takes brakes and wears the HMD at the beginning of each session. A subject typically complete four sessions within $45-60$ minutes in total. First two sessions follow the usual routine from Table \ref{table:procedure} with a task of choosing a predefined target object from multiple choices (varying randomly from three to eight choices). At each task, a target object appears at a new random location chosen from uniformly distributed sample locations. Total number of target selection task in a single session is 50. Definition of success at a task is being able to choose the target object by gazing at the target object at least for 3 seconds. On the other hand, definition of a failure case can be expressed as follows: If our system fails to detect the case of gazing correctly, user continuously gaze at the target, and push a button to give a feedback to the system. System uses this as an online training data, and takes advantage of the new data for the upcoming tasks within a session. Third and fourth sessions follows the same routine as in first two sessions. However, those sessions does not go through the calibration phase. A third session is initiated with the calibration data captured from the same user during a second session. A fourth session is initiated with the captured data from an another subject during an another session in the past. Figure \ref{fig:statistics} summarizes results of our early extended experiments. 

According to our early investigation in Table \ref{fig:statistics}, task of choosing by gazing at a target is likely to be more successful with a calibration phase at the beginning of a session. We also observed that system can be initiated with a calibration data from a previous session with lower success rations. Thus, requiring further online training during usage. At the extreme, our findings suggest that it is possible to initiate the system with a calibration data from a different person with least amount of success rations and maximum amount of online training during usage.

\textbf{Sampling Rates and latency:} Given our current hardware and uC, we are able to sample up to $250~Hz$ with our first prototype coupled to a mini computer (Raspberry Pi model B). On the other hand, our second prototype coupled to a personal computer samples at $100~Hz$. A better uC with a dedicated ADC for each analogue pin can sample all channels simultaneously, thus such a uC can increase sampling rates up to $kHz$ range.
 We observe that the smartphone's USB does not provide a reliable serial communication at $250~Hz$. Thus, we  intentionally use only a $100~Hz$ sampling rates with our first prototype. Our GPR implementation can easily run at above $100~Hz$ even on the smartphone (our datasets all contain less than $100$ calibration points). 


\section{Discussion}
\label{section:discussion}

We believe that gaze tracking is a critical feature that will contribute to the success of HMDs in the consumer market. We propose a new sensor technology that enables gaze tracking at low cost and in a small form factor, ideal for the HMD use case.

\textbf{Feedback from subjects:} In informal feedback, subjects found our system to be accurate for the given tasks shown during a subjective test. 

One common problem reported by the subjects was sudden jumps in gaze estimation after and during a blink. Similarly, slightly closing an eye lid was found to cause a shift in the gaze estimation. In the future work we plan to address these problems by detecting blinks from the LED inputs as well as predicting the eye lid status. Some subjects experienced lower spatial accuracy in certain regions within the screen. We believe this to be due to differences in eye sizes, relative eye locations, and eye distances. Simply adding more channels to each eye should help to address this problem. Finally, the initial calibration phase with a four-by-four grid was found to be cumbersome, and we will continue to investigate on how to further decrease the number of calibration points. Self-calibration through saliency maps~\cite{sugano2015self} may be an option.

\textbf{Future Work:} Kalman filtering is a common method to enhance accuracy (but may add latency), as are visual anchors \cite{Kumar:2008:IAG:1344471.1344488}. Expanding the subject pool can also help us in clustering subjects for estimating a good initial calibration data for each person. We will evaluate these methods in the near future. An analogy from coded aperture optics is an inspiring direction for our future work: merging our technique with backlights of Spatial Light Modulators (SLMs) could potentially provide an interesting all-in-one solution for sensing and illumination, similar to bi-directional displays~\cite{hirsch2009bidi,SDTP:SDTP1540}.

Our subject pool contains subjects with and without eye prescriptions. During a subjective test, subjects with eye prescriptions were asked to take off their glasses and wear contact lenses instead. If our technique was to be used in a non-HMD scenario, corrective glasses would need to be supported.

In a near future, VR users may also benefit from wearable facial recognition techniques using photo-reflective sensors \cite{masai2015affectivewear} or from vital sign monitoring, such as heart-rate and blood oxygenation, which have been estimated remotely using photodiodes \cite{cennini2010heart} in the past. A similar methodology using LEDs could be explored through simple changes in our circuitry. Thus, those methodologies may allow us to improve in accuracy by considering effects of facial changes due to mood, respiration or heart-rate. Saccadic reaction times (SRTs) are well explored by the medical community as a basis for
 health condition prediction techniques~\cite{mazumdar2014comparison}. Our current prototype can also be trained to detect SRTs (yellow regions in Figure~\ref{fig:accuracy}) in the future. Rogers et al.~\cite{rogers2015approach} recently demonstrated that blinks and head movements captured using infrared, accelerometer, and gyroscope sensors in a HMD can be used for the task of user identification. All these sensors are readily available in our prototypes. User identification dimension can be added to our demonstration by merging \cite{rogers2015approach}'s findings with our proposal.

\section{Conclusion}
\label{section:conclusion}

Gaze sensing HMDs promise to provide a dataset for a set of useful tasks: foveated rendering, predicting a user's mood and health, predicting visual saliency at a personal level, and providing a VR experience that is unique, that is a natural fit for you. 

We describe a novel gaze tracker methodology using only LEDs. Our proposal's main contribution is exploiting the sensing capability of LEDs with a GPR implementation using Minkowski distances inside a HMD system. We provide a detailed description of our algorithms and our hardware, which we believe is the simplest gaze-detecting design to date. We have shown two unique prototypes that validates our proposal in the context of a traditional off-the-shelf HMD. As gaze-sensing HMDs are poised to enter the market, we hope to inspire more research on simple, low-power non-imaging approaches such as ours.

\section*{Acknowledgements}
We thank the reviewers for insightful feedback, and also Duygu Ceylan, Ward Lopes, Fu-Chung Huang, Joohwan Kim, and Orazio Gallo for the fruitful discussions, and also to our subjects for participating to our subjective experiments.

\bibliographystyle{unsrt}
\bibliography{references}

\begin{thebibliography}{10}

\bibitem{jacob2003eye}
RJ~Jacob and Keith~S Karn.
\newblock Eye tracking in human-computer interaction and usability research:
  Ready to deliver the promises.
\newblock {\em Mind}, 2(3):4, 2003.

\bibitem{majaranta2014eye}
P{\"a}ivi Majaranta and Andreas Bulling.
\newblock Eye tracking and eye-based human--computer interaction.
\newblock In {\em Advances in Physiological Computing}, pages 39--65. 2014.

\bibitem{san2010gaze}
Javier San~Agustin, John~Paulin Hansen, and Martin Tall.
\newblock Gaze-based interaction with public displays using off-the-shelf
  components.
\newblock In {\em ACM International Conference Adjunct Papers on Ubiquitous
  Computing}, pages 377--378, 2010.

\bibitem{stellmach2012look}
Sophie Stellmach and Raimund Dachselt.
\newblock Look \& touch: gaze-supported target acquisition.
\newblock In {\em ACM CHI}, pages 2981--2990, 2012.

\bibitem{turner2013eye}
Jayson Turner, Jason Alexander, Andreas Bulling, Dominik Schmidt, and Hans
  Gellersen.
\newblock Eye pull, eye push: Moving objects between large screens and personal
  devices with gaze and touch.
\newblock In {\em Human-Computer Interaction -- INTERACT 2013}, pages 170--186.
  2013.

\bibitem{turner2014cross}
Jayson Turner, Andreas Bulling, Jason Alexander, and Hans Gellersen.
\newblock Cross-device gaze-supported point-to-point content transfer.
\newblock In {\em Symp. on Eye Tracking Research and Appl.}, pages 19--26,
  2014.

\bibitem{san2009low}
Javier San~Agustin, Henrik Skovsgaard, John~Paulin Hansen, and Dan~Witzner
  Hansen.
\newblock Low-cost gaze interaction: ready to deliver the promises.
\newblock In {\em ACM CHI}, pages 4453--4458, 2009.

\bibitem{san2010evaluation}
Javier San~Agustin, Henrik Skovsgaard, Emilie Mollenbach, Maria Barret, Martin
  Tall, Dan~Witzner Hansen, and John~Paulin Hansen.
\newblock Evaluation of a low-cost open-source gaze tracker.
\newblock In {\em Symposium on Eye-Tracking Research \& Applications}, pages
  77--80, 2010.

\bibitem{johansen2011low}
Sune~Alstrup Johansen, Javier San~Agustin, Henrik Skovsgaard, John~Paulin
  Hansen, and Martin Tall.
\newblock Low-cost vs. high-end eye tracking for usability testing.
\newblock In {\em ACM CHI}, pages 1177--1182, 2011.

\bibitem{mantiuk2012yourself}
Rados{\l}aw Mantiuk, Micha{\l} Kowalik, Adam Nowosielski, and Bartosz Bazyluk.
\newblock {\em Do-it-yourself eye tracker: Low-cost pupil-based eye tracker for
  computer graphics applications}.
\newblock Springer, 2012.

\bibitem{dietz2003very}
Paul Dietz, William Yerazunis, and Darren Leigh.
\newblock Very low-cost sensing and communication using bidirectional {LEDs}.
\newblock In {\em UbiComp}, pages 175--191, 2003.

\bibitem{round1907note}
Henry~J Round.
\newblock A note on carborundum.
\newblock {\em Electrical World}, 49(6):309, 1907.

\bibitem{holmqvist2012eye}
Kenneth Holmqvist, Marcus Nystr{\"o}m, and Fiona Mulvey.
\newblock Eye tracker data quality: what it is and how to measure it.
\newblock In {\em Symp. on Eye Tracking Research and Applications}, pages
  45--52, 2012.

\bibitem{6852086}
G.~Corbellini, K.~Aksit, S.~Schmid, S.~Mangold, and T.~Gross.
\newblock Connecting networks of toys and smartphones with visible light
  communication.
\newblock {\em IEEE Communications Magazine}, 52(7):72--78, 2014.

\bibitem{schmid2014visible}
Stefan Schmid, Josef Ziegler, Thomas~R Gross, Manuela Hitz, Afroditi Psarra,
  Giorgio Corbellini, and Stefan Mangold.
\newblock (in)visible light communication: combining illumination and
  communication.
\newblock In {\em ACM SIGGRAPH 2014 Emerging Technologies}, page~13, 2014.

\bibitem{lange2011multicolour}
V~Lange, F~Lima, and D~K{\"u}hlke.
\newblock Multicolour led in luminescence sensing application.
\newblock {\em Sensors and Actuators A: Physical}, 169(1):43--48, 2011.

\bibitem{ben2008led}
Moshe Ben-Ezra, Jiaping Wang, Bennett Wilburn, Xiaoyang Li, and Le~Ma.
\newblock An led-only brdf measurement device.
\newblock In {\em IEEE Conference on Computer Vision and Pattern Recognition},
  pages 1--8, 2008.

\bibitem{li2014color}
Shuai Li and Ashish Pandharipande.
\newblock Color sensing and illumination with led lamps.
\newblock In {\em IEEE International Conference on Consumer Electronics}, pages
  1--2, 2014.

\bibitem{li2015led}
Shuai Li and Ashish Pandharipande.
\newblock {LED}-based color sensing and control.
\newblock {\em IEEE Sensors Journal}, 15(11):6116--6124, 2015.

\bibitem{li2015human}
Tianxing Li, Chuankai An, Zhao Tian, Andrew~T Campbell, and Xia Zhou.
\newblock Human sensing using visible light communication.
\newblock In {\em Int. Conf. on Mobile Computing and Networking}, pages
  331--344, 2015.

\bibitem{an2015visible}
Chuankai An, Tianxing Li, Zhao Tian, Andrew~T Campbell, and Xia Zhou.
\newblock Visible light knows who you are.
\newblock In {\em International Workshop on Visible Light Communications
  Systems}, pages 39--44, 2015.

\bibitem{robinson1963method}
David Robinson et~al.
\newblock A method of measuring eye movemnent using a scieral search coil in a
  magnetic field.
\newblock {\em Bio-medical Electronics, IEEE Transactions on}, 10(4):137--145,
  1963.

\bibitem{marg1951development}
Elwin Marg.
\newblock Development of electro-oculography: Standing potential of the eye in
  registration of eye movement.
\newblock {\em AMA Archives of Ophthalmology}, 45(2):169--185, 1951.

\bibitem{mayberry2014ishadow}
Addison Mayberry, Pan Hu, Benjamin Marlin, Christopher Salthouse, and Deepak
  Ganesan.
\newblock ishadow: Design of a wearable, real-time mobile gaze tracker.
\newblock In {\em MobiSys}, pages 82--94. ACM, 2014.

\bibitem{SDTP:SDTP1540}
Uwe Vogel, Daniel Kreye, Bernd Richter, Gerd Bunk, Sven Reckziegel, Rigo
  Herold, Michael Scholles, Michael Törker, Christiane Grillberger, Jörg
  Amelung, Sven-Thomas Graupner, Sebastian Pannasch, Michael Heubner, and
  Boris~Mitrofanovich Velichkovsky.
\newblock 8.2: Bi-directional {OLED} microdisplay for interactive {HMD}.
\newblock {\em SID Symposium Digest of Technical Papers}, 39(1):81--84, 2008.

\bibitem{naugle2013two}
Etienne Naugle and Reynald Hoskinson.
\newblock Two gaze-detection methods for power reduction in near-to eye
  displays for wearable computing.
\newblock In {\em Int. Conf. on Wireless and Mobile Comp., Net. and Comm.},
  pages 675--680, 2013.

\bibitem{topal2014low}
Cihan Topal, Serkan Gunal, Onur Ko{\c{c}}deviren, Atakan Dogan, and Omer~Nezih
  Gerek.
\newblock A low-computational approach on gaze estimation with eye touch
  system.
\newblock {\em IEEE Transactions on Cybernetics}, 44(2):228--239, 2014.

\bibitem{grover2005eye}
D~Grover, T~Delbruck, and M~King.
\newblock An eye tracking system using multiple near-infrared channels with
  special application to efficient eyebased communication.
\newblock In {\em European Conference on Eye Movement}, pages 36--39, 2005.

\bibitem{Rasmussen06gaussianprocesses}
Carl~Edward Rasmussen.
\newblock {\em Gaussian processes for machine learning}.
\newblock MIT Press, 2006.

\bibitem{6467271}
F.~Martinez, A.~Carbone, and E.~Pissaloux.
\newblock Gaze estimation using local features and non-linear regression.
\newblock In {\em IEEE Int. Conf. on Image Processing}, pages 1961--1964, 2012.

\bibitem{noris2008calibration}
Basilio Noris, Karim Benmachiche, and Aude Billard.
\newblock Calibration-free eye gaze direction detection with gaussian
  processes.
\newblock In {\em VISAPP}, pages 611--616, 2008.

\bibitem{sewell2010real}
Weston Sewell and Oleg Komogortsev.
\newblock Real-time eye gaze tracking with an unmodified commodity webcam
  employing a neural network.
\newblock In {\em ACM CHI}, pages 3739--3744, 2010.

\bibitem{kunka2009non}
Bartosz Kunka and Bozena Kostek.
\newblock Non-intrusive infrared-free eye tracking method.
\newblock In {\em Signal Processing Algorithms, Architectures, Arrangements,
  and Applications}, pages 105--109, 2009.

\bibitem{van1997near}
Thomas~JTP van~den Berg and Henk Spekreijse.
\newblock Near infrared light absorption in the human eye media.
\newblock {\em Vision Research}, 37(2):249--253, 1997.

\bibitem{boucouvalas1996iec}
AC~Boucouvalas.
\newblock {IEC} 825-1 eye safety classification of some consumer electronic
  products.
\newblock {\em IEE Colloquium on Optical Free Space Communication Links}, pages
  13:1--13:6, 1996.

\bibitem{nguyen2002differences}
Karlene Nguyen, Cindy Wagner, David Koons, and Myron Flickner.
\newblock Differences in the infrared bright pupil response of human eyes.
\newblock In {\em Symp. on Eye Tracking Research \& Appl.}, pages 133--138,
  2002.

\bibitem{bray1957ordination}
J~Roger Bray and John~T Curtis.
\newblock An ordination of the upland forest communities of southern
  {Wisconsin}.
\newblock {\em Ecological Monographs}, 27(4):325--349, 1957.

\bibitem{lance1966computer}
GN~Lance and WT~Williams.
\newblock Computer programs for hierarchical polythetic classification
  (“similarity analyses”).
\newblock {\em The Computer Journal}, 9(1):60--64, 1966.

\bibitem{mazumdar2014comparison}
Deepmala Mazumdar, JJ~M Pel, Manish Panday, Rashima Asokan, Lingam Vijaya,
  B~Shantha, Ronnie George, and J~Van Der~Steen.
\newblock Comparison of saccadic reaction time between normal and glaucoma
  using an eye movement perimeter.
\newblock {\em Indian journal of ophthalmology}, 62(1):55, 2014.

\bibitem{sugano2015self}
Yusuke Sugano and Andreas Bulling.
\newblock Self-calibrating head-mounted eye trackers using egocentric visual
  saliency.
\newblock In {\em ACM Symposium on User Interface Software \& Technology},
  pages 363--372, 2015.

\bibitem{Kumar:2008:IAG:1344471.1344488}
Manu Kumar, Jeff Klingner, Rohan Puranik, Terry Winograd, and Andreas Paepcke.
\newblock Improving the accuracy of gaze input for interaction.
\newblock In {\em Symp. on Eye Tracking Research Appl.}, pages 65--68, 2008.

\bibitem{hirsch2009bidi}
Matthew Hirsch, Douglas Lanman, Henry Holtzman, and Ramesh Raskar.
\newblock {BiDi} screen: a thin, depth-sensing {LCD} for {3D} interaction using
  light fields.
\newblock {\em ACM Transactions on Graphics (TOG)}, 28(5), 2009.

\bibitem{masai2015affectivewear}
Katsutoshi Masai, Yuta Sugiura, Katsuhiro Suzuki, Sho Shimamura, Kai Kunze,
  Masa Ogata, Masahiko Inami, and Maki Sugimoto.
\newblock Affectivewear: towards recognizing affect in real life.
\newblock In {\em Proceedings of the 2015 ACM International Joint Conference on
  Pervasive and Ubiquitous Computing and Proceedings of the 2015 ACM
  International Symposium on Wearable Computers}, pages 357--360. ACM, 2015.

\bibitem{cennini2010heart}
Giovanni Cennini, Jeremie Arguel, Kaan Ak{\c{s}}it, and Arno van Leest.
\newblock Heart rate monitoring via remote photoplethysmography with motion
  artifacts reduction.
\newblock {\em Optics Express}, 18(5):4867--4875, 2010.

\bibitem{rogers2015approach}
Cynthia~E Rogers, Alexander~W Witt, Alexander~D Solomon, and Krishna~K
  Venkatasubramanian.
\newblock An approach for user identification for head-mounted displays.
\newblock In {\em Proceedings of the 2015 ACM International Symposium on
  Wearable Computers}, pages 143--146, 2015.

\end{thebibliography}
\end{document}